\pdfoutput=1

\documentclass[11pt]{article}

\usepackage[preprint]{acl}

\usepackage{times}
\usepackage{latexsym}
\usepackage{booktabs}
\usepackage{multirow, makecell}

\usepackage{centernot}
\usepackage{marvosym}
\usepackage{etoolbox}
\usepackage{cuted}
\usepackage{hyperref}
\AfterEndEnvironment{strip}{\leavevmode}

\usepackage{fdsymbol}
\usepackage[T1]{fontenc}
\usepackage[utf8]{inputenc}
\usepackage{microtype}
\usepackage{inconsolata}

\usepackage{graphicx}
\usepackage{tabularx}
\usepackage{algpseudocode}
\usepackage{microtype}
\usepackage{subcaption}
\usepackage{algorithm}
\usepackage{algpseudocode}
\usepackage{multirow}
\usepackage{multicol}
\usepackage{rotating}
\usepackage{colortbl}

\usepackage{amsmath}
\DeclareMathOperator*{\argmax}{\textit{\textbf{argmax}}}

\usepackage{xspace}

\usepackage{enumitem} 
\usepackage{adjustbox}

\usepackage{comment}

\definecolor{dkblue}{rgb}{0,0,0.5}

\newcommand{\bfund}[1]{\underline{\textbf{#1}}}

\newcommand{\wicd}{{WiCkeD}}
\newcommand{\wicds}{\wicd\space}
\newcommand{\nota}{\textit{None of the above}}

\title{WiCkeD: A Simple Method to Make Multiple Choice Benchmarks More Challenging}

\author{Ahmed Elhady$^{1}$ \quad Eneko Agirre$^{1}$ \quad Mikel Artetxe$^{1,2}$ \\
$^{1}$HiTZ Center, University of the Basque Country (UPV/EHU) \qquad $^{2}$Reka AI \\
\texttt{\{ahmed.salemmohamed,e.agirre,mikel.artetxe\}@ehu.eus} }

\begin{document}
\maketitle

\begin{abstract}
    We introduce {\wicd}, a simple method to increase the complexity of existing multiple-choice benchmarks by randomly replacing a choice with \textit{"None of the above"}, a method often used in educational tests. We show that \wicds can be automatically applied to any existing benchmark, making it more challenging. We apply \wicds to 6 popular benchmarks and use it to evaluate 18 open-weight LLMs. The performance of the models drops {12.1} points on average with respect to the original versions of the datasets. When using chain-of-thought on 3 MMLU datasets, the performance drop for the \wicds variant is similar to the one observed when using the LLMs directly, showing that \wicds is also challenging for models with enhanced reasoning abilities.
    \wicds also uncovers that some models are more sensitive to the extra reasoning required, providing additional information with respect to the original benchmarks. We relase our code and data at
    \url{https://github.com/ahmedselhady/wicked-benchmarks}.
\end{abstract}

\begin{figure*}[th!]
    \centering
    \includegraphics[width=0.7\linewidth]{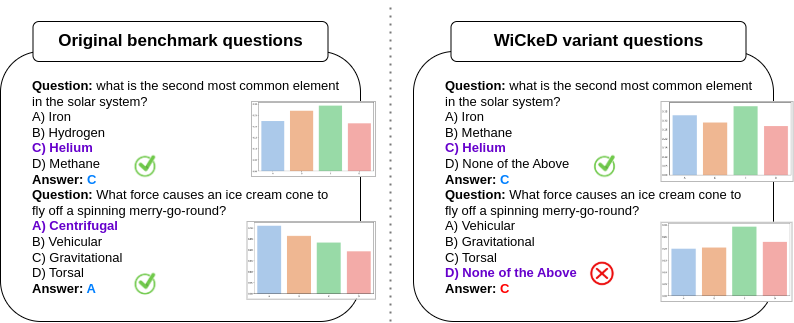}
    \caption{Two samples from MMLU-Pro (left) and its \wicds variant (right), where \textit{Hydrogen} and \textit{Centrifugal} were removed. Correct answers in bold. Llama-3.1 8B correctly answers both original questions but fails on the \wicds variant for the second question. The probability distribution of the model for each answer is also shown.}
    \label{fig:wicked}
\end{figure*}

\section{Introduction}

Multiple choice question (MCQ) benchmarks are widely used to evaluate Large
Language Models (LLMs). This format consists of a question and a limited set of
options, which include a correct (or best) answer and several distractors that
are either incorrect or less appropriate (see Figure~\ref{fig:wicked}). There
are various MCQ datasets that focus on different capabilities, including
factual knowledge and reasoning as in MMLU
\cite{hendrycks2021measuringmassivemultitasklanguage} and Arc-challenge
\cite{clark2018thinksolvedquestionanswering}, common sense as in Commonsense-QA
\cite{talmor-etal-2019-commonsenseqa}, truthfulness as in TruthfulQA
\cite{lin2022truthfulqameasuringmodelsmimic}, and domain-specific knowledge
\cite{ALONSO2024102938, hosseini2024benchmarklongformmedicalquestion}.
Unfortunately, most of these benchmarks got quickly saturated in the recent era
dominated by LLMs, motivating harder datasets to better gauge the abilities of
newer models. However, developing benchmarks is a laborious and expensive
process.

Motivated by this, several recent works have explored strategies to make
existing benchmarks harder, which can serve as an alternative to creating new
benchmarks from scratch. For example, \citet{gema2024mmlu} identified erroneous
questions in the MMLU benchmark, and re-annotated 3k questions to be harder and
more robust. Similarly, \citet{wang2024mmluprorobustchallengingmultitask}
presented MMLU-Pro, a harder version of the MMLU benchmark that replaces noisy
questions with harder ones and expands the number of distractors to include
more plausible yet incorrect ones. While increasing the number of distractors
reduces the probability of correct guesses by chance, creating plausible and
coherent distractors is challenging and often requires manual verification
\citep{mcintosh2024inadequacieslargelanguagemodel}.

In this work, we propose a simple yet effective method to make existing
benchmarks more challenging without the need to add distractors. Namely, we
present the \bfund{Wi}ld-\bfund{C}ard \bfund{D}istractor (\wicd) which creates
a variant of any existing MCQ benchmark by keeping the question unchanged, and
randomly replacing one of the choices with a wild-card distractor, \nota~ (see
Figure~\ref{fig:wicked}). We create \wicd~variants of 6 popular benchmarks, and
use them to evaluate 18 open-weight LLMs varying in size, model family, and
training recipe. The \wicds datasets suffer a performance drop of 7.2-19.7
points with respect to the original datasets, depending on the model being
evaluated. Using chain-of-thought does not prevent the drop (1.4-14.6), showing
that \wicds can be used to assess reasoning capabilities. The large variance
across models shows that \wicds is not only challenging, but it also uncovers
differences in model capabilities that are not captured by the original
benchmarks.

\section{Related Work}

\subsection{Challenges in LLMs MCQ Benchmarks }

Several works raised concerns about the effectiveness of MCQ benchmarks in LLM
assessment. For example, \citet{balepur2024artifactsabductionllmsanswer} showed
that some LLMs can answer MCQs using only the answer choices, without seeing
the questions, and perform well-above baselines. Furthermore, more works
suggested that LLMs are biased towards certain answer keys (A/B/C/D) due to
unbalanced prior probabilities rather than actual knowledge
\citep{myrzakhan2024openllmleaderboardmultichoiceopenstylequestions,clark2018thinksolvedquestionanswering}.
Another line of research attributes LLMs hallucinations to being unable to
identify when they lack sufficient knowledge about the subject matter
\citep{li2024thinktwicetrustingselfdetection, ji-etal-2022-answer}.
Nonetheless, current evaluation benchmarks do not assess this capability
effectively. We view our work as an addition towards efficient evaluation of
LLMs to avoid spurious correlations and account for knowledge and reasoning
gaps.

\subsection{None of the Above in Educational Tests}

Multiple-choice questions (MCQs) are effective assessments when they include
plausible distractors, as they encourage deeper processing to think not only
about why a given choice is correct, but also why other choices are wrong and
improve knowledge recall \cite{little2019role, little2015optimizing}. The use
of \nota~ as a distractor in MCQs is an area of research and debate. It can
provide unique insight into the understanding of the examinees and potentially
differentiate their abilities \cite{DiBattista03042014, dochy2001assessment}.
However, \nota~ can affect the confidence of the examinee, leading them to
avoid selecting \nota~ as the correct answer, even when it is true
\cite{little2023does, odegard2007none}. Nevertheless, incorporating \nota~ into
practice tests can enhance the learning process by encouraging deeper
engagement with the material \cite{DiBattista03042014,
    pezeshkpour-hruschka-2024-large, zheng2024largelanguagemodelsrobust}.

\section{Methodology}

We propose a method to automatically create a more challenging version of any
existing MCQ benchmark without requiring any manual annotation. The difficulty
of MCQ has been linked to the reasoning necessary to discriminate between
competing options \citep{mcintosh2024inadequacieslargelanguagemodel,
    wang2024mmluprorobustchallengingmultitask}. We hypothesize that detecting the
absence of the correct answer within the provided options is more challenging
than selecting the correct one. To that end, we propose to add a wild-card
choice \nota. Note that adding \nota~ as an additional option would not make
sense, as the correct answer is always the correct option, we thus propose to
replace one of the options instead.

\subsection{The \wicds Algorithm}

Given a benchmark that consists of $M$ examples where each has $N$ choices (one
correct answer and $N-1$ distractors), we uniformly sample one option to be
omitted, and append the wildcard option \nota. When the correct option is
replaced, the new correct option is \nota. When a distractor option is
replaced, the correct option continues to be correct. Figure~\ref{fig:wicked}
shows the result of applying \wicds to two examples. The goal is to produce a
variant for each benchmark that contains the same number of $M$ examples.

\begin{figure}[t!]
    \centering
    \includegraphics[width=1.\columnwidth]{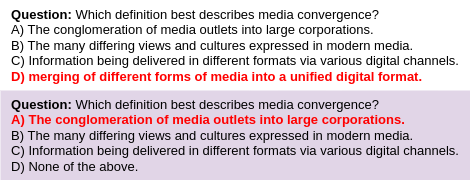}
    \caption{Applying \wicds on a single best answer (SBA) example (best answer D, second best answer A) would lead to an incoherent \wicds variant (incorrectly having \nota ~as the gold correct answer instead of A). We thus copy SBA examples verbatim, see \S~\ref{sec:coherence} for details.  }
    \label{fig:sba-vs-sca}
\end{figure}

\subsection{Coherence of \wicds Examples}
\label{sec:coherence}
The above algorithm does not always produce coherent examples. In some cases, there are more than one correct candidate, but only one of them is the most appropriate (see Figure \ref{fig:sba-vs-sca}, where D is the best answer and A is the second best answer). With the above procedure, when the replaced option is the correct one (e.g. option D in the figure), the \wicd ~variant would add \nota~and take this option as the correct one. However, this would be incoherent, because having removed D, A becomes the next best option. We call these examples Single Best Answer (SBA) as opposed to Single Correct Answer (SCA, where the distractors are all incorrect).  As we want to keep the same number of examples  we avoid adding \nota ~to SBA examples and copy them unchanged to the \wicd ~variant of the benchmark.

In order to train an example classifier to detect SBA examples, we selected
four representative benchmarks (MMLU, MMLU-Pro, Truthful-QA and
Commonsense-QA), sampled 4000 examples, and split them into evaluation (25\%)
and train (75\%). We used GPT-4o-mini to automatically label the examples as
SBA or SCA, and further annotated the evaluation split manually. Given the cost
and slow speed of GPT-4o-mini, we used the synthetic labels to train a
classifier based on BERT\footnote{\url{https://huggingface.co/ahmedselhady/bert-base-uncased-sba-clf}}
\cite{devlin2019bertpretrainingdeepbidirectional}.

The recall on SBA examples for the classifier is over 98.9\%, showing that we
are able to detect nearly all SBA examples, and would thus have 1.1\% noisy
\wicds examples (that is, examples in the benchmark that have \nota ~as the
correct option even if a correct option exists). See Appendix
~\ref{app:annotation-alignment} for more details about the training and
evaluation procedure.

\section{Experimental Setup}

\subsection{Benchmarks}

We apply \wicds to six popular MCQ benchmarks that assess the knowledge,
language comprehension, reasoning, and truthfulness of LLMs: MMLU, MMLU-Pro,
MMLU-Redux, CommonsenseQA, Truthful-QA, and Arc-challenge. To ensure
reproducibility, we use Eval-Harness \cite{eval-harness}. Given that the
selection of the option to be replaced is random, we generate five \wicd
~variants for each benchmark, and report mean and standard deviation. 

Regarding the amount of SBA examples, MMLU, MMLU-Redux and MMLU-pro have the
largest amount ($\sim20\%$), with the rest of the benchmarks having less than
5\% (see Appendix \ref{app:annotation-alignment}). SBA examples are copied
verbatim to the \wicd ~variants, but the fact that at least 80\% of the
examples are effectively altered makes the \wicd ~variants significantly more
challenging, as we will see. Other benchmarks have less than 5\% SBAs; we also
leave them unchanged.

\subsection{Models}

\begin{table}[t!]
    \renewcommand{\arraystretch}{0.9}
    \begin{adjustbox}{max width=1.\columnwidth}

        \begin{tabular}{lrcrrr}
            \toprule
            \textbf{Model}             & \textbf{Size} & \makecell{\textbf{IT}} & \textbf{Original} & \textbf{\wicd} & \multicolumn{1}{c}{$\mathbf{\Delta}$}                        \\
            \midrule
            DS-R1-Llama                & 8B            & -                      & 56.6              & 48.6           & -7.9  {\scriptsize $\pm 1.1\%$}                              \\
            DS-R1-Qwen                 & 7B            & -                      & 60.8              & 53.4           & -7.3 {\scriptsize $\pm 1.6\%$}                               \\
            \midrule
            \multirow{4}{*}{Llama-3.1} & 8B            & -                      & 61.4              & 52.2           & -9.2 {\scriptsize $\pm 1.7\%$}                               \\
                                       & 8B            & $\checkmark$           & 66.0              & 55.0           & -11.0  {\scriptsize $\pm 0.9\%$}                             \\
                                       & 70B           & -                      & 76.8              & 67.0           & -9.8 {\scriptsize $\pm 2.1\%$}                               \\
                                       & 70B           & $\checkmark$           & 77.1              & 64.5           & -12.6 {\scriptsize $\pm 1.3\%$}                              \\
            \midrule
            \multirow{2}{*}{Mistral}   & 7B            & -                      & 59.8              & 46.5           & -13.2 {\scriptsize $\pm 1.2\%$}                              \\
                                       & 7B            & $\checkmark $          & 59.0              & 47.2           & -11.8 {\scriptsize $\pm 1.1\%$}                              \\
            \midrule
            \multirow{6}{*}{Qwen-2.5}  & 7B            & -                      & 74.7              & 54.9           & -19.7 {\scriptsize $\pm 1.5\%$}                              \\
                                       & 7B            & $\checkmark $          & 73.5              & 59.0           & -14.5  {\scriptsize $\pm 1.3\%$}                             \\
                                       & 14B           & -                      & 78.9              & 66.3           & -12.6 {\scriptsize $\pm 2.1\%$}                              \\
                                       & 14B           & $\checkmark $          & 78.9              & 66.6           & -12.3  {\scriptsize $\pm 1.8\%$}                             \\
                                       & 72B           & -                      & 84.6              & 72.6           & -12.0 {\scriptsize $\pm 0.9\%$}                              \\
                                       & 72B           & $\checkmark $          & 82.6              & 69.3           & -13.3 {\scriptsize $\pm 1.0\%$}                              \\
            \midrule
            \multirow{4}{*}{Gemma-2}   & 9B            & -                      & 67.3              & 56.3           & -10.9 {\scriptsize $\pm 1.2\%$}                              \\
                                       & 9B            & $\checkmark $          & 73.3              & 57.6           & -15.7  {\scriptsize $\pm 1.2\%$}                             \\
                                       & 27B           & -                      & 68.0              & 54.6           & -13.4 {\scriptsize $\pm 2.0\%$}                              \\
                                       & 27B           & $\checkmark $          & 74.8              & 61.9           & -12.9 {\scriptsize $\pm 2.3\%$}                              \\
            \bottomrule
            \textbf{Average}           &               &                        & \textbf{70.78}    & \textbf{58.52} & \cellcolor{red!25} \textbf{-12.2} {\scriptsize $\pm 26.3\%$} \\
            \toprule
        \end{tabular}
    \end{adjustbox}
    \caption{Average performance on original and \wicds variants of the six benchmarks. IT: instruction-tuned. $\mathbf{\Delta}$: degradation from original performance}
    \label{tab:detailed-results}

\end{table}

We evaluate \wicds on 18 open-weight models covering different families and
sizes. Namely, we evaluate the base and instruction-tuned models of Qwen2.5 7B,
14B and 72B \citep{qwen2025qwen25technicalreport}, Llama3.1 8B and 70B
\citep{grattafiori2024llama3herdmodels}, Gemma2 9B and 27B
\citep{gemmateam2024gemma2improvingopen}, and Mistral-7B
\citep{jiang2023mistral7b}. We also selected two DeepSeek-R1 models for their
improved reasoning capabilities: distill-Lllama3.1-8B and distill-Qwen7
\citep{deepseekai2025deepseekr1incentivizingreasoningcapability}.

The LLM models are evaluated on the benchmarks following the standard
multiple-choice prompting procedure
\cite{robinson2023leveraginglargelanguagemodels}, see Appendix \ref{app:mcp}.
We set the number of few-shot examples to five, in order to ensure that in most
cases there is at least one example where {\nota} is the correct option.

In addition, we also evaluate the LLM models using zero-shot chain-of-thoughts
prompting (CoT) on the three benchmarks commonly used to assess the reasoning
capabilities of LLMs: MMLU, MMLU-Pro, and MMLU-Redux. We set the maximum
generation length to 4096, unless limited by the model itself.

\section{Results and Discussion}

\subsection{Main Results}
\label{sec:main-res}
Table~\ref{tab:detailed-results} shows the mean accuracy of the models on the original and \wicds benchmarks, with a significant drop in performance. Qwen2.5-7B suffers the largest degradation (19.73\%), while its DeepSeek-R1 distilled version (DeepSeek-R1-Qwen7B) suffers the least (7.35\%). This suggests that models with better reasoning capabilities, like R1, are better equipped to deal with the added complexity.

\begin{table}[t!]
    \renewcommand{\arraystretch}{0.9}
    \begin{adjustbox}{max width=1.\columnwidth}

        \begin{tabular}{lrccrrcr}
            \toprule
                                       &                   &                        & \multicolumn{2}{c}{\textbf{Direct}} &                                   & \multicolumn{2}{c}{\textbf{CoT}}                                                               \\
            \cmidrule{4-5} \cmidrule{7-8}
            \textbf{Model}             & \textbf{Size}     & \makecell{\textbf{IT}}
                                       & \textbf{\wicd}
                                       & $\mathbf{\Delta}$
                                       &                   & \textbf{\wicd}         & $\mathbf{\Delta}$                                                                                                                                                        \\
            \midrule
            DS-R1-Llama                & 8B                & -                      & 30.3                                & -4.1                              &                                  & 80.1                    & -2.0                              \\
            DS-R1-Qwen                 & 7B                & -                      & 30.6                                & -4.3                              &                                  & 74.9                    & -2.5                              \\
            \midrule
            \multirow{2}{*}{Llama-3.1} & 8B                & -                      & 39.7                                & -3.2                              &                                  & 53.9                    & -5.8                              \\
                                       & 8B                & $\checkmark$           & 43.6                                & -2.7                              &                                  & 57.2                    & -3.4                              \\
            \midrule
            \multirow{2}{*}{Mistral}   & 7B                & -                      & 35.9                                & -3.4                              &                                  & 36.3                    & -11.62                            \\
                                       & 7B                & $\checkmark $          & 33.5                                & -5.7                              &                                  & 43.8                    & -4.97                             \\
            \midrule
            \multirow{4}{*}{Qwen-2.5}  & 7B                & -                      & 45.5                                & -6.9                              &                                  & 43.0                    & -14.62                            \\
                                       & 7B                & $\checkmark $          & 47.1                                & -5.3                              &                                  & 55.4                    & -1.73                             \\
                                       & 14B               & -                      & 55.6                                & -3.6                              &                                  & 61.5                    & -3.97                             \\
                                       & 14B               & $\checkmark $          & 56.7                                & -3.4                              &                                  & 64.0                    & -1.43                             \\
            \midrule
            \multirow{4}{*}{Gemma-2}   & 9B                & -                      & 36.1                                & -12.2                             &                                  & 41.2                    & -8.93                             \\
                                       & 9B                & $\checkmark $          & 44.1                                & -9.3                              &                                  & 56.3                    & -4.36                             \\
                                       & 27B               & -                      & 36.1                                & -10.8                             &                                  & 59.2                    & -4.07                             \\
                                       & 27B               & $\checkmark $          & 51.3                                & -3.8                              &                                  & 60.3                    & -3.77                             \\
            \bottomrule
            \textbf{Average}           &                   &                        & \textbf{41.86}                      & \cellcolor{red!25} \textbf{-5.62} &                                  & \textbf{\textbf{56.26}} & \cellcolor{red!25} \textbf{-5.24} \\
            \toprule
        \end{tabular}

    \end{adjustbox}
    \caption{Performance on \wicds variants for MMLU, MMLU-pro, and MMLU-Redux without and with CoT.  IT: instruction-tuned. $\mathbf{\Delta}$: degradation from the original benchmark.}
    \label{tab:detailed-results-cot}

\end{table}

Prominently, the \wicds variants shuffle the ranking of models. For example,
the Qwen2.5-7B and Qwen2.5-7B-IT models originally performed close to the
Llama-3.1-70B model. However, on the \wicds variants, they lag behind it by
13\% and 8\%, respectively. Similar patterns can be seen in Gemma-2-9B-IT and
Gemma-2-27B-IT, which lag behind Llama-3.1-70B by 9.5\% and 5.3\%,
respectively. Qwen2.5-72B and Llama-3.1-70B are the models that perform best in
\wicd. There is no clear advantage from instruction-tuning, as results vary
depending on the model family.

\subsection{Chain of Thought Results}

Table~\ref{tab:detailed-results-cot} shows the performance of the
models\footnote{Due to computing constraints we could not run CoT for the
    $\sim$70B models} on the MMLU, MMLU-pro, and MMLU-Redux \wicds benchmarks. The
drop for these three benchmarks without CoT (direct columns in the table) is
lower than the other three benchmarks, but applying CoT does not reduce the
drop in \wicds variants, which stays above 5\%. This is remarkable given that
CoT is very effective at improving results on MMLU and related benchmarks.
Instruction-tuned models experience significantly less degradation than their
base models, especially when using CoT (see
Appendix~\ref{app:instruct-vs-base-cot} for additional details). Notably, the
DeepSpeed-R1 distilled models, Qwen7B and Llama3.1-8B, suffer ~2\% each.
Similarly, Instruction tuned Qwen2.5 7B and 14B suffer less than 2\%. We
hypothesize this is due to their enhanced reasoning capabilities.

\section{Conclusion}

In this paper, we introduced a simple automatic method to create more
challenging variants from existing MCQ benchmark. The large drop in the results
shows that \wicds challenges the knowledge and reasoning of LLMs, as they need
to identify the absence of the correct answer, even when using CoT. We showed
that models with better reasoning capabilities suffer less in \wicd, such as
the original Qwen7B and its distilled version of DS-R1. We see \wicds as an
addition towards efficient evaluation of LLMs to avoid spurious correlations
and challenge reasoning and knowledge gaps. A deeper look into why some models
are more sensitive to \wicds than others can provide significant insights about
uncovered limitations. We release all the code and data under open licenses.

\section*{Limitations}

We manually confirmed the applicability of \wicds on some popular
multiple-choice benchmarks whose questions can be categorised into SBAs and
SCAs. However, for other benchmarks, \wicds might need further verification.
Furthermore, we focus the evaluation of \wicds on open-weight LLMs only, the
performance of closed models, such as GPT-4 and Claude, has not been explored
yet.

\section*{Acknowledgments}

\bibliography{custom}
\newpage
\appendix

\begin{table*}[th!]
    \centering
    \begin{adjustbox}{max width=1.0\linewidth}
        \begin{tabular}{rrrrrrr}
            \toprule
                           & \textbf{MMLU} & \textbf{MMLU-Pro} & \textbf{TQA} & \textbf{CSQA} & \textbf{Recall} & \textbf{Precision} \\
            \midrule
            Manual         & 17.3          & 12.3              & 3.3          & 3.8           & --              & --                 \\
            GPT-4o-mini    & 18.2          & 13                & 4.2          & 3.9           & 98.5            & 97.4               \\
            SBA Classifier & 19.6          & 14.2              & 4.5          & 4             & 98.9            & 95.1               \\
            \bottomrule
        \end{tabular}
    \end{adjustbox}
    \caption{The Percentage of Single Best Answer (SBA) questions in 1K questions sampled uniformly from  MMLU, MMLU-Pro, TruthfulQA (TQA), and CommonsenseQA (CSQA) as determined by our manual Annotations, GPT-4o-mini, and our trained SBA classifier. Recall and precision are computed with respect to the manual annotation.
    }
    \label{tab:detailed-sba-percentages}
\end{table*}

\begin{table*}[th!]
    \centering
    \begin{adjustbox}{max width=1.\linewidth}
        \begin{tabular}{|l|}
            \toprule
            \texttt{"A single correct answer question is a question that can have exactly one correct answer from a given set of choices.} \\
            \texttt{A single best answer question can have a most appropriate answer (for example, if this answer is omitted, another answer will be correct).
            }                                                                                                                              \\
            \texttt{Classify the following questions into SBA and non-SBA questions. Assign a label of 1 if the question is a SBA question and a label of 0 otherwise.
            }                                                                                                                              \\
            \texttt{Question: \{question\}
                Class:"
            }                                                                                                                              \\ \hline
        \end{tabular}
    \end{adjustbox}
    \caption{SBA Annotation Prompt Template}
    \label{tab:prompt_template}
\end{table*}

\section{Detecting Single Best Answer examples}
\label{app:annotation-alignment}

To ensure the reliability of the automatic identification of single-best-answer
(SBA) questions, we uniformly sample 4K questions from the MMLU, MMLU-Pro,
Commonsense-QA, and Truthful-QA benchmarks, which we divide into 1K and 3K
splits. We then manually annotate the 1K samples and optimize GPT-4o-mini
prompt on them for best recall. Table ~\ref{tab:prompt_template} shows the
prompt template for GPT-4o-mini, which we used to annotate the 4K questions.
The 3K split was then used to train our Bert-based SBA classifier on them. The
classifier was trained for 2 epochs, using a learning rate of 1e-04. The model
was frozen, except for the last layer and the classification head.

\begin{table*}[t!]
    \begin{adjustbox}{max width=1.0\linewidth}
        \begin{tabular}{rrrrrr}
            \toprule
            \textbf{MMLU} & \textbf{MMLU-Pro} & \textbf{MMLU-Redux} & \textbf{TruthfulQA} & \textbf{Commonsense QA} & \textbf{Arc Challenge} \\
            \midrule
            20.3\%        & 16.8\%            & 14.7\%              & 3.2\%               & 3.7\%                   & 5.2\%                  \\
            \bottomrule
        \end{tabular}
    \end{adjustbox}
    \caption{The Percentage of Single Best Answer (SBA) questions in the benchmarks as determined by our SBA classifier. We do not apply \wicd\space to SBA questions as it can break their coherence.
    }
    \label{tab:sba-percentages}
\end{table*}

Table ~\ref{tab:detailed-sba-percentages} shows the percentages of SBA
questions on the 1K split as determined by our manual annotations, GPT-4o-mini,
and the SBA classifier. The classifier is the preferred one, as it is the most
conservative, that is, it detects the most SBA examples, which would be copied
verbatim to the \wicds variant of the benchmark. The evaluation figures in the
table confirm this choice, as the classifier has higher recall. The small drop
in precision is harmless, as it means that we will not add \nota ~option to
those examples, and will be copied verbatim. In other words, we can estimate
that \wicds contains 1\% of incoherent examples (where there is a valid option
even if \nota ~is recorded as the correct option), and 5\% of examples which do
not have a \nota~ option even if we could have added it if the classifier had
100\% precision. These figures confirm the high quality of the \wicds variants.
Table ~\ref{tab:sba-percentages} shows the final SBA percentages for each
benchmark as determined by the classifier.

\begin{figure}[th!]
    \centering
    \includegraphics[width=1.\columnwidth]{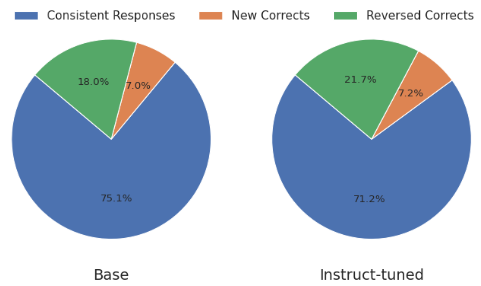}
    \caption{The changes in models' answers of the original benchmarks and the \wicds variant using chain-of-thoughts. }
    \label{fig:piechart}
\end{figure}

\section{Instruct vs Base Models on Chain of Thought}

\label{app:instruct-vs-base-cot}
Results of CoT suggest the instruct models experience less degradation than their base models. To better understand why this happens, we analyze their answers. Figure~\ref{fig:piechart} shows the change in answers from the original to the \wicds variants. Instruction-tuned models are less prone to reverse correct answers and can correct original mistakes in \wicd. This suggests that \wicds is useful for better gauging the reasoning capabilities of the models.

\section{Multiple Choice Prompting}
\label{app:mcp}

In multiple choice prompting, the model is prompted with few-shot
demonstrations \textit{c} and a question \textit{q} and the set of choices $ A
    = \{A,B,C,D\}$. It generates a probability of the answer label $ a \epsilon A$
conditioned on the prefix prompt given by:
\begin{equation}
    \renewcommand{\arraystretch}{0.4} %
    \label{eqn:choice-prob}
    \textsc{P}(a|c,q) = \prod_{t=1}^{T} p(a_t|c,q<T)
\end{equation}

The model's answer is set to:
\begin{equation}
    \renewcommand{\arraystretch}{0.4} %
    \label{eqn:icl-prob}
    \argmax_{a\epsilon A}(P(a| c,q))
\end{equation}

\begin{figure}[th!]
    \centering
    \includegraphics[width=1\columnwidth]{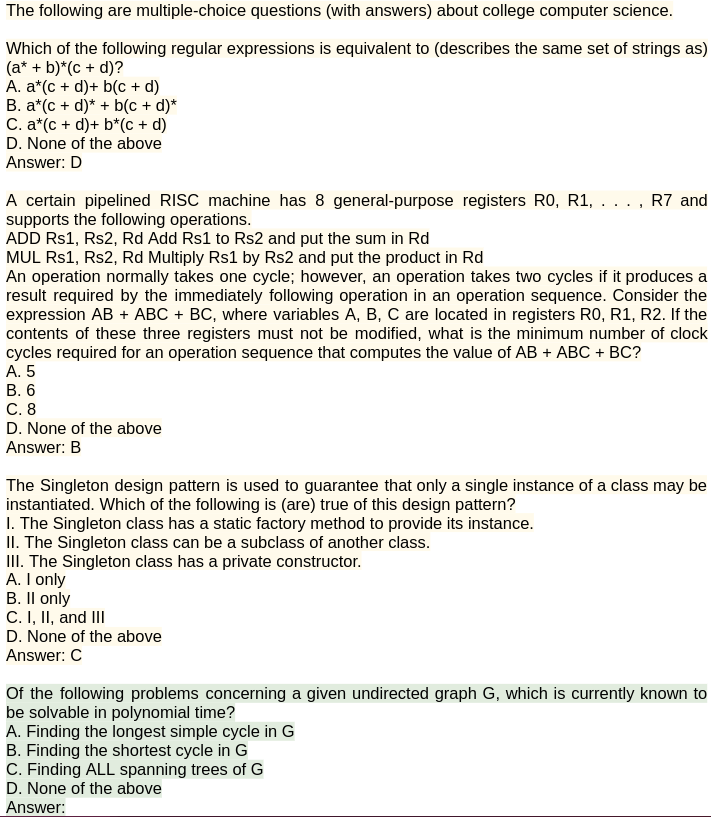}
    \caption{Examples from the MMLU computer science task using \wicd.  We show 3-shot for brevity, but 5-shot was actually used in the experiments for the main results.}
    \label{fig:mmlu-cs-examples}
\end{figure}

\begin{figure}[th!]
    \centering
    \includegraphics[width=1\columnwidth]{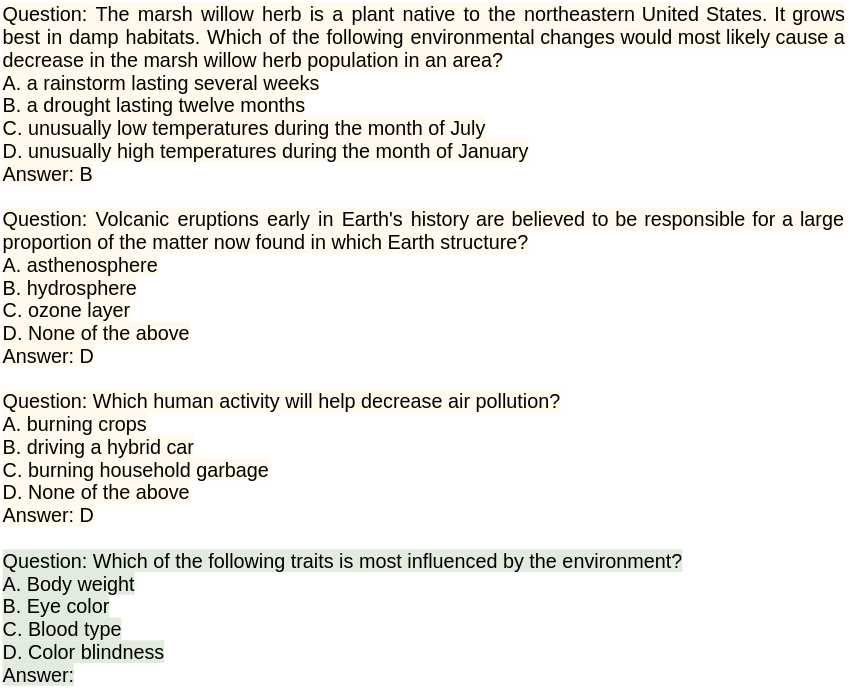}
    \caption{Examples from the AllenAi Arc challenge using \wicd. We show 3-shot for brevity, but 5-shot was actually used in the experiments for the main results. The first few-shot example does not include \nota~ option because it was classified as SBA question.}
    \label{fig:arc-examples}
\end{figure}

\begin{figure}[th!]
    \centering
    \includegraphics[width=1\columnwidth]{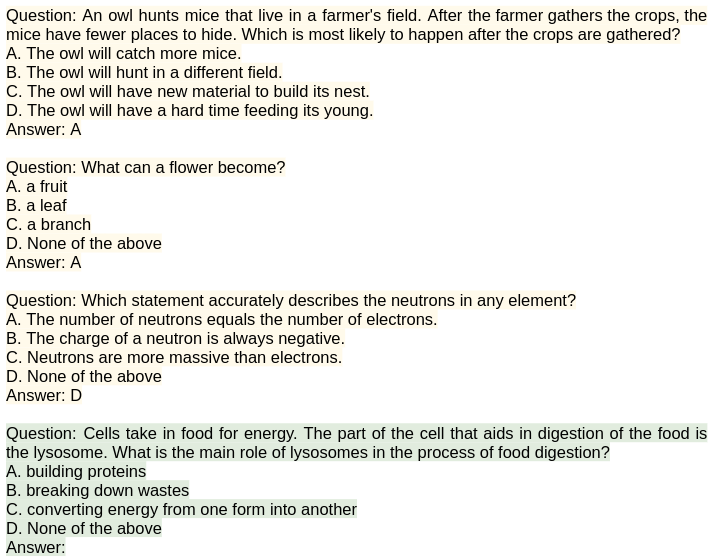}
    \caption{Examples from the Common-sense QA using \wicd.  We show 3-shot for brevity, but 5-shot was actually used in the experiments for the main results. The first few-shot example does not include \nota~ option because it was classified as SBA question. }
    \label{fig:csqa-examples}
\end{figure}

\begin{figure}[th!]
    \centering
    \includegraphics[width=1\columnwidth]{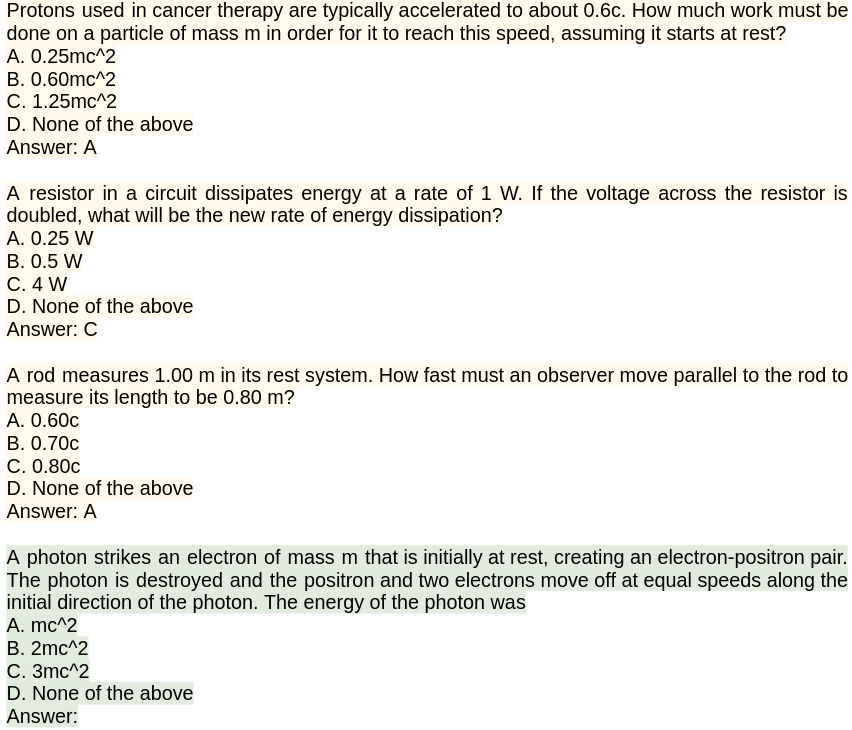}
    \caption{Examples from the MMLU-Redux using \wicd. We show 3-shot for brevity, but 5-shot was actually used in the experiments for the main results. }
    \label{fig:mmlur-examples}
\end{figure}

Figures ~\ref{fig:mmlu-cs-examples}, ~\ref{fig:arc-examples},
~\ref{fig:csqa-examples}, and ~\ref{fig:mmlur-examples} show example prompts
for the MMLU college computer science, Arc Challenge, Common-sense QA, and
MMLU-Redux benchmarks, respectively.

\end{document}